# Data-Centric AI Paradigm Based on Application-Driven Fine-Grained Dataset Design

Huan Hu
ChinaUnicom
huh30@chinaunicom.cn

Yajie Cui
ChinaUnicom
cuiyj62@chinaunicom.cn

Zhaoxiang Liu*
ChinaUnicom
liuzx178@chinaunicom.cn

Shiguo Lian*
ChinaUnicom
liansg@chinaunicom.cn

## Abstract

*Deep learning has a wide range of applications in industrial scenario, but reducing false alarm (FA) remains a major difficulty. Optimizing network architecture or network parameters is used to tackle this challenge in academic circles, while ignoring the essential characteristics of data in application scenarios, which often results in increased FA in new scenarios. In this paper, we propose a novel paradigm for fine-grained design of datasets, driven by industrial applications. We flexibly select positive and negative sample sets according to the essential features of the data and application requirements, add the remaining samples to the training set as uncertain sample sets, and finally get the sample sets with at least three classes. Taking mask-wearing detection as example, we collect more than 10,000 mask-wearing recognition samples covering various application scenarios as our experimental data. Experimental results show that compared with traditional dataset design our proposed method achieves better results and effectively reduces FA. This dataset design method may be applied to various practical applications and is expected to be a new data-centric AI paradigm. We make all contributions available to the research community for broader use. The contributions will be available at https://github.com/huh30/OpenDatasets.*

## 1. Introduction

With the development of deep learning [1-2], computer vision algorithms have been widely used in industrial scenario. To achieve better results, it is necessary to

---
*corresponding author

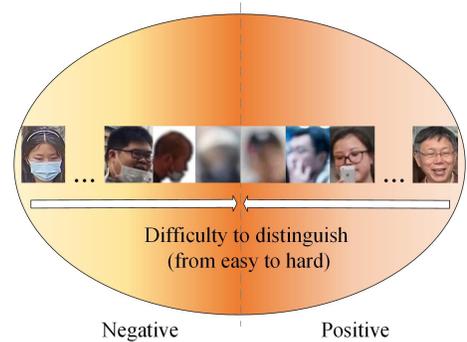

Figure 1. The feature paradigm of mask-wearing recognition dataset, with distinguishable samples at two ends and indistinguishable samples in the centre. Difficulty level of recognition gradually increases from the two ends to the center, and the samples closer to the center are more likely to be misidentified.

provide large-scale relevant datasets. However, obtaining a significant amount of industrial scenario data is exceedingly challenging. In academic circles, pre-training on large-scale open-source datasets still reigns supreme. To accomplish associated tasks, better networks and parameters are then fine-tuned on smaller real-world scenario datasets. But the approach may bring some problems, of which we take object detection task as example for explanation. VOC [3] and COCO [4] are the most commonly used open-source datasets for model pre-training in the field of object detection. But for industrial application scenarios, the data division of VOC and COCO is too coarse. Taking person counting task as an example, COCO labels parts of human body, such as fingers or single leg, as person. In complex industrial scenarios, such kind of data division method can easily lead to objects similar to legs or fingers being falsely

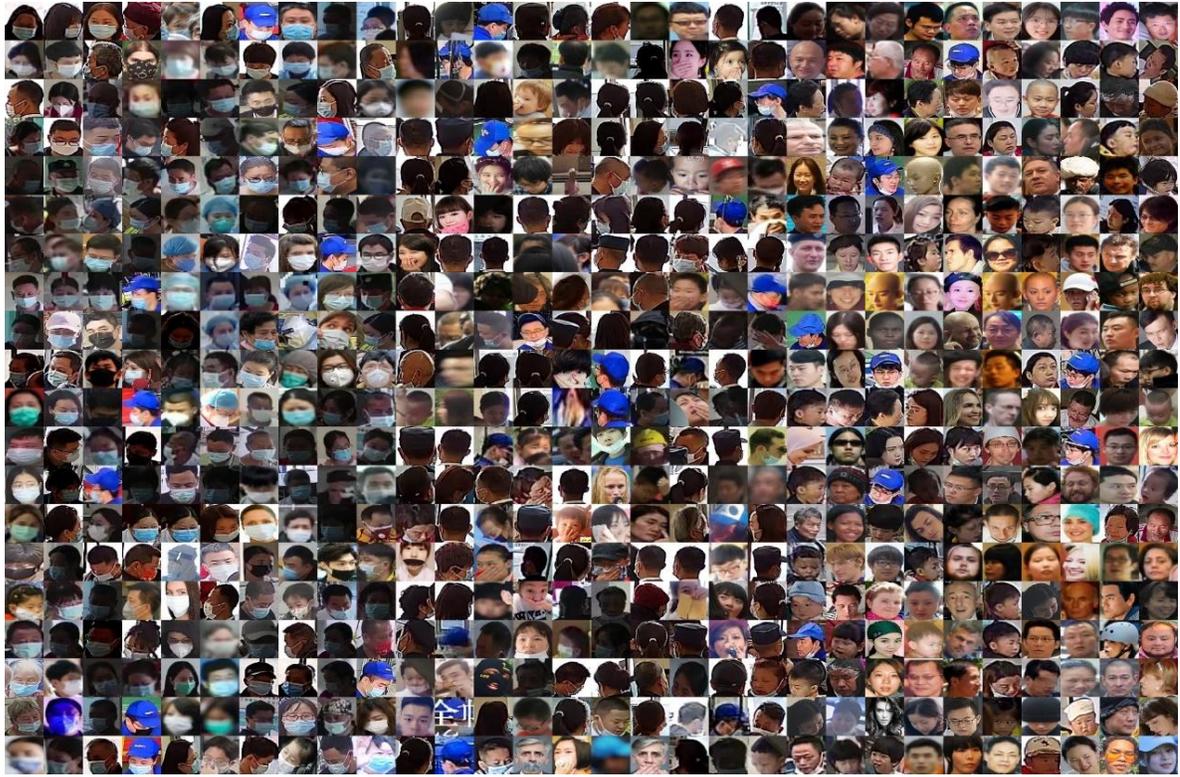

Figure 2. Sample images from our collected datasets.

detected as human, resulting in incorrect counts.

Taking facemask-wearing recognition for example, Figure 1 shows the feature paradigm of mask-wearing recognition dataset, with distinguishable samples at two ends and indistinguishable samples in the centre. Difficulty level of recognition gradually increases from the two ends to the center, and the samples closer to the center are more likely to be misidentified. Many datasets for industrial applications fit this paradigm as well, such as helmet detection, work uniform recognition, glove wearing recognition, etc. In industrial application scenarios, we designate the categories that need extra attention as positive samples, such as not wearing mask in mask recognition, not wearing helmet in helmet detection, not wearing gloves in glove recognition, etc. Positive samples are closely related to management and security in practical applications, and precision rate is more important than recall rate, thus, we need to ensure the recognition precision of positive samples. When the model misidentifies negative samples as positive ones, false alarm occurs. Frequent false alarms will bring very terrible user experience. Therefore, reducing false alarms is a big challenge we need to address.

Current mainstream dataset design methods ignore the characteristics of the data itself and the actual scenarios. Take mask-wearing recognition as an example, according to the mainstream approach of dataset design, the data will be conservatively assigned to positive and negative sample sets. When the data scale is not large enough, the model trained in this way is disastrous. Because the actual application scenarios are very complex, i.e. irregularly wearing (e.g., a part of the mouth or nose is exposed), large angles and large poses, low quality, small targets, etc. model trained on these samples may produce a large number of false alarm (FA). Here, FA stands for misidentifying mask-wearing face or other non-face object as not wearing mask. It is difficult to solve this problem by just updating the AI model such as an object detection model or an image classification model. The dataset should be finely designed according to the actual application scenarios and the own characteristics of dataset itself.

In this paper, we propose the method that finely designs the dataset driven by applications. By mining the own features of data and flexibly selecting the positive and negative sample sets according to the actual requirements, and adding the remaining samples to the training set as uncertainty sample sets, the AI model with at least three classes (positive, negative and uncertainty) can be trained on the dataset which significantly reduces FA.

The contributions of this paper are the following two aspects.

(1) We propose an application-driven paradigm for fine-grained design of datasets with mask-wearing

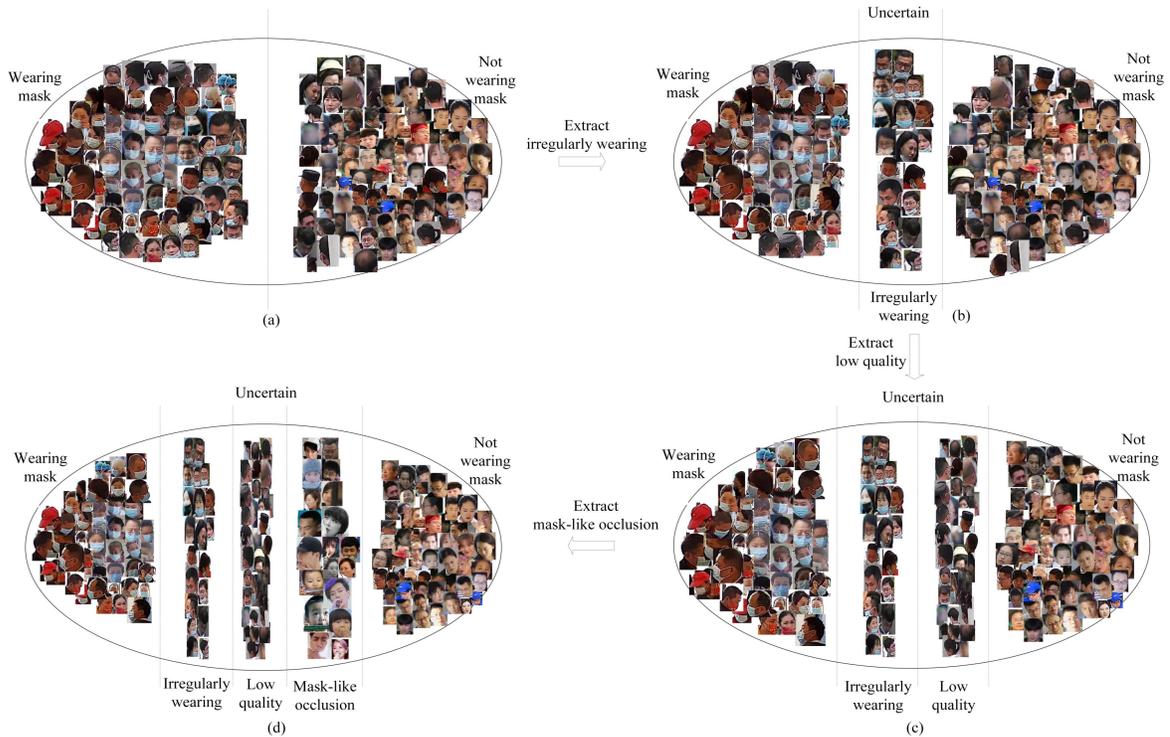

Figure 3. The architecture of our framework. Construct uncertain category datasets based on data attributes and application requirements. (a) Straightforward method of dataset design. In (b) we extract the samples with irregular wearing from (a), the low-quality samples are extracted in (c), and we extract the samples with mask-like occlusion in (d).

recognition as example.

(2) We show the proposed dataset design method's effectiveness with experiments, and open source our finely-designed mask-wearing recognition dataset.

The rest of the paper is arranged as follows. Some related work is introduced in Section 2. In Section 3, the proposed dataset fine-design method is presented in detail. The experiments are done and results are given in Section 4. Finally, in Section 5, the conclusion is drawn.

## 2. Related Work

Large-scale datasets with high-quality annotations play a crucial role in driving better computer vision models. For image classification, the most commonly used datasets are ImageNet [5-6]. ImageNet is a large image dataset built to facilitate the development of image recognition technology, containing 14,197,122 images and 21,841 categories. For object detection, PASCAL VOC is an early benchmark that contains 17,000 images in 20 categories. Then there is COCO in 2014, which is currently the most widely adopted benchmark for object detection. It contains 118,000 images and 860,000 instance annotations in 80 categories.

ImageNet and COCO datasets with the characteristics of large-scale and high-quality together with deep learning have revolutionized the face of computer vision(CV). Most new algorithms assess their performance on these datasets and provide pre-trained models. We can reuse these pre-trained models in similar domains through transfer learning [7]. The advantage of transfer learning is that it can produce excellent results with little in the way of training time, training data, or computational resources. Therefore, transfer learning is often the first choice in research and industry when developing deep learning-based models for CV tasks. However, pre-trained models may carry inherent hazards for domains with unique business needs. Data division of open-source datasets is coarse and inaccurate for actual application scenarios, which makes the transfer-learned model incapable of distinguishing difficult samples in real scenarios.

Ng proposes data-centric AI [8] to address the above problems by designing targeted subsets of data. Data-centric AI is the discipline of systematically engineering the data needed to successfully build an AI system. For many practical applications, it is more efficient to improve the data than the network structure. Meanwhile, since many businesses simply lack enormous data volumes, the focus has to shift from big data to good data.

We illustrate our data design approach with mask-wearing recognition task. COVID-19 has made a huge impact on our lives. Wearing mask is the most effective way to protect against COVID-19. AIZOO [9] provides the face mask detection dataset which includes parts of WIDER Face [10] with Masked Faces (MAFA) datasets [11]. In AIZOO, training set contains 6,120 images and 13,593 faces, and the test set has 1830 images and 5082 faces. The Moxa 3K dataset created by Roy et al. [12] is used for training and evaluating face mask detection models, which consists of 3,000 images with different scenarios from close-up faces to crowded scenes, using 2,800 images for training and 200 images for testing. However, the above mentioned datasets are only simply divided into two categories: masked and unmasked, without considering the characteristics of the data and the needs of the application scenarios, leading to a lot of FA. In our work, we adopt the data-centric AI and carefully design the mask-wearing dataset, which effectively reduces the FA of the model.

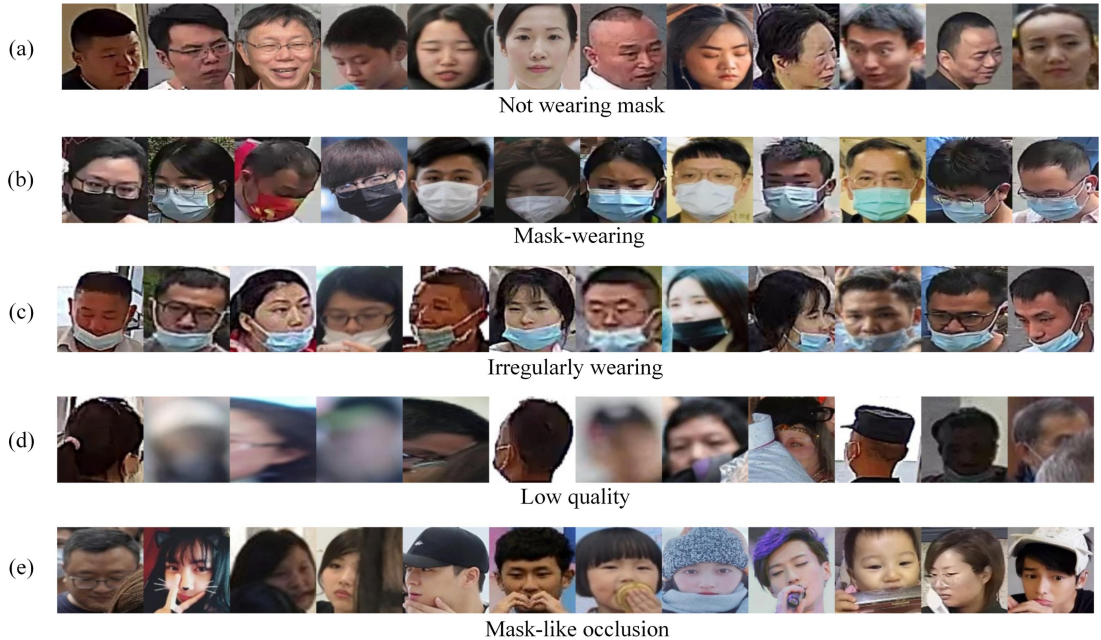

Figure 4. Example of all category images. (a) Typical samples of not wearing mask. (b) Representative samples of mask-wearing. (c) Irregularly wearing: people wear a mask but part of their nose or mouth is exposed. (d) Low Quality: occlusion, overexposure, blurring and so on make it impossible to confirm whether a person is wearing a mask or not. (e) Mask-like occlusion: mouth or nose is subtle occluded by non-mask objects.

## 3. The Proposed Dataset Fine-Design Method

In this section, we introduce the proposed approach with the example of mask-wearing recognition, including two parts, i.e., data collection and dataset design.

### 3.1. Data collection

We collect data from 75 real application scenarios including hospitals, health clinics, neighborhoods, schools, streets, offices, gas stations, factories, cold chains, warehouses, etc., and extract head parts in each scenario by existing head detectors, and finally build more than 10,000 images of mask-wearing recognition data. The dataset covers both masked and unmasked human head images with various angles, illumination, size, and quality. It is manually labeled by a number of people in numerous rounds, and it is divided into positive and negative subsets based on the mainstream dataset construction, as shown in Figure 2.

### 3.2. Dataset design

The overall architecture of our dataset design scheme is presented in Figure 3. The currently mainstream dataset division method is described in Figure 3(a), which conservatively divides the dataset into positive and negative categories. Figure 3(b) extracts the samples with irregular wearing from Figure 3(a), Figure 3(c) extracts the low-quality samples, and Figure 3(d) extracts the samples with mask-like occlusion. In the proposed dataset designing scheme, we select the unambiguously non-masked and masked samples as positive and negative sets as depicted in Figure 4(a) and Figure 4(b), which is based on the characteristics of the data and the requirements of the application scenario. The samples that are easy to cause FA, such as irregular wearing, low quality, mask-like occlusion, etc., are selected as the uncertain class to form the final training set.

### 3.2.1. Straightforward dataset design

Straightforward dataset design refers to directly dividing the dataset into positive and negative categories, as shown in Figure 3(a), which is the current mainstream method. In mask-wearing recognition, due to the indistinguishable samples such as irregularly wearing, low quality, mask-like occlusion etc., model trained on the dataset designed in a straightforward manner are disastrous and will produce a large number of FA.

### 3.2.2. Dataset design considering of irregularly wearing (IW)

We define irregularly mask-wearing as shown in Figure 4(c), in which people wear a mask but part of their nose or mouth is exposed. These samples are extremely similar to mask-wearing which should be categorized as not wearing mask. If we train model on the dataset taking irregularly mask-wearing as positive sample, some images of correctly mask-wearing will be mistakenly identified as not wearing mask, resulting in FA. Therefore, we extract this kind of data separately and classify it as uncertain category, as shown in Figure 3(b).

### 3.2.3. Dataset design considering of low quality (LQ)

Low quality data means that it is hard to identify whether a person is wearing mask or not due to shooting angle, illumination, distance, occlusion, etc. As shown in Figure 4(d), factors like occlusion, overexposure, blurring, only the back of people's head or part of the side face can be seen and so on make it impossible to confirm whether a person is wearing a mask or not. Therefore, the above data needs to be delt with separately and classified as the uncertainty category, as illustrated in Figure 3(c).

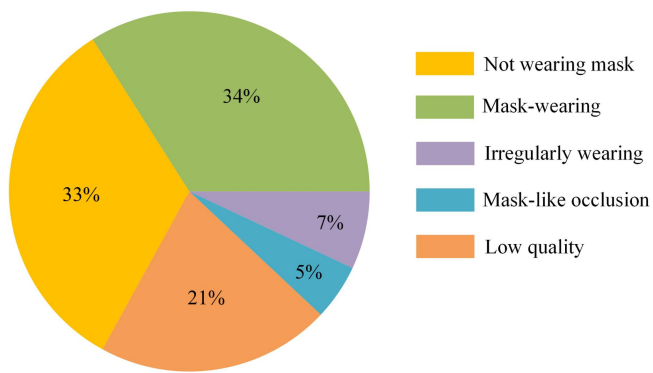

Figure 5. Distribution of training dataset. The dataset includes the five kinds of faces: without mask wearing, with mask wearing, with mask irregularly wearing, with mask-like occlusion, and with low quality.

### 3.2.4. Dataset design considering of mask-like occlusion (MLO)

Mask-like occlusion data refers to the mouth or nose being subtle occlusion by non-mask objects as illustrated in Figure 4(e). Mask-like occlusion data is classified as negative samples according to the traditional data design method, which may lead to incorrect recognition of mask-wearing samples similar to MLO. However, extracting MLO data will decrease the recall of not wearing mask. There are some actual requirements that are less stringent for recall rate of not wearing mask yet have high expectations for precision rate, because frequent FA bring very bad user experience. Hence, we need to further separate MLO data out and then put them to the uncertainty category as shown in Figure 3(d).

## 4. Experiments

In this section, we conduct experiments on a collected mask-wearing recognition dataset containing more than 10,000 images for training and 1,000 images for testing which do not overlap the training set. We use resnet50 [13] as backbone and softmax loss as loss function. We fine-tuned model on ImageNet's pre-trained model, with training learning rate being 1e-3, cosine decay applied, optimizer Adam [14], and we train for totally 50 epochs.

According to our design, the training dataset is divided into five parts as shown in Figure 5, of which 3384 are not-wearing-mask, 3465 are mask-wearing, 587 are irregularly wearing, 2319 are low-quality, and the remaining 375 are mask-like occlusion. We unify the irregularly wearing, low-quality, and mask-like occlusion into the uncertainty class for training.

We demonstrate the effectiveness of our dataset design method through ablation experiments. Original in Table 1 is the dataset obtained according to the straightforward dataset design which strictly divides training dataset into two categories: wearing a mask and not wearing a mask. In order to verify the impact of IW, LQ and MLO on FA, we extract IW, LQ and MLO classes from Original, and construct uncertain classes by combining IW, LQ and MLO classes, design seven training datasets on which we train resnet50 classification models. Finally, we test and compare the performance of the trained models on the same test dataset.

The results of our experiments are described in Table 1. Specifically, when the uncertainty class contains any one of IW, LQ and MLO, the FAR (false alarm rate) decreases. This indicates that IW, LQ and MLO all have an impact on FA, and it can be seen from the table that LQ has the greatest impact on FA. When the uncertainty class contains any two of IW, LQ and MLO, the FAR decreases more. When the uncertainty class contains IW, LQ and MLO all, the FAR reaches the minimum value of 0.8%, which indicates that our dataset design method can greatly reduce

FA by introducing uncertain classes according to application scenarios.

| Dataset Design | FAR |
|---|---|
| Original | 8.9% |
| Extract IW only | 4.1% |
| Extract LQ only | **3.8%** |
| Extract MLO only | 5.8% |
| Extract IW+LQ | 1.4% |
| Extract IW+MLO | 3.6% |
| Extract LQ+MLO | 3.3% |
| Extract IW+LQ+MLO | **0.8%** |

Table 1. Results on our designed dataset. With the same basic AI model, various dataset design methods are tested to compare the corresponding FARs.

## 5. Conclusion

In this paper, we propose a dataset fine-design method to solve such AI landing problem as high false alarm rate in practical applications. To keep the scalability, the data samples are collected from various practical scenarios. Then, the dataset is constructed by finely classifying the samples into positive sample set, negative sample set and uncertain sample sets according to data characteristics and performance requirements on e.g. FAR. With the fine-designed dataset, the AI model such as object detection or image classification is trained. Taking facemask-wearing recognition as example, more than 10,000 samples are collected and finely-designed for various datasets. Additionally, the datasets are open-source and available publicly. Experimental results show that the AI models trained by the datasets designed with our proposed method achieve better recognition results and effectively reduce FA, compared with traditional dataset design. Additionally, this dataset fine-design method can be extended to various scenarios of industrial applications. Therefore, it is expected to be a general data-centric AI paradigm based on fine-design of dataset for obtain good performance in practical applications.